# Scaling Neuro-symbolic Problem Solving: Solver-Free Learning of Constraints and Objectives

MARIANNE DEFRESNE*, LAAS-CNRS, Université de Toulouse, CNRS, INSA, Toulouse, France
ROMAIN GAMBARDELLA, Télécom-Paris, France
SOPHIE BARBE, TBI, Université de Toulouse, CNRS, INRAE, INSA, ANITI, France
THOMAS SCHIEX*, Université de Toulouse, ANITI, INRAE, UR 875, France

**Background:** In the ongoing quest for hybridizing discrete reasoning with neural nets, there is an increasing interest in neural architectures that can learn how to solve discrete reasoning or optimisation problems from natural inputs, a task that Large Language Models seem to struggle with.
**Objectives:** We introduce a differentiable neuro-symbolic architecture and a loss function dedicated to learning how to solve NP-hard reasoning problems.
**Methods:** Our new probabilistic loss allows for learning both the constraints and the objective – possibly non-linear – of a combinatorial problem. Thus, it delivers a complete model that can be scrutinized and completed with side constraints. By pushing the combinatorial solver out of the training loop, our architecture also offers scalable training while exact inference gives access to maximum accuracy.
**Results:** We empirically show that it can efficiently learn how to solve NP-hard reasoning problems from natural inputs. On three variants of the Sudoku benchmark – symbolic, visual, and many-solution –, our approach requires a fraction of data and training time of other hybrid methods. On a visual Min-Cut/Max-cut task, it optimizes the regret as well as a Decision-Focused-Learning regret-dedicated loss. Finally, it efficiently learns the energy optimisation formulation of the large real-world problem of designing proteins.

## 1 Introduction

In recent years, several hybrid neural architectures have been proposed that integrate discrete reasoning or optimisation within neural network architectures. The motivations for these works are varied. For discrete optimisation and reasoning experts, the main motivation is to deal with uncertainty proactively. Available data is leveraged to produce a solution (decision), whose quality approaches that achievable with a full knowledge of the

*Corresponding Authors.

Authors' Contact Information: Marianne Defresne, ORCID: 0000-0001-7043-9260, marianne.defresne@laas.fr, LAAS-CNRS, Université de Toulouse, CNRS, INSA, Toulouse, Toulouse, , France; Romain Gambardella, romain.gambardella@telecom-paris.fr, Télécom-Paris, 91120 Palaiseau, , France; Sophie Barbe, ORCID: 0000-0003-2581-5022, sophie.barbe@insa-toulouse.fr, TBI, Université de Toulouse, CNRS, INRAE, INSA, ANITI, 31077 Toulouse, , France; Thomas Schiex, ORCID: 0000-0001-6049-3415, thomas.schiex@inrae.fr, Université de Toulouse, ANITI, INRAE, UR 875, 31326 Toulouse, , France.

problem to solve. This motivation has led to the introduction of Decision-Focused Learning (DFL) (Mandi et al. 2024). Such hybrid architectures are also of interest for users who have a dataset of discrete structured objects of interest (sequences, trees, graphs,...) with associated correlated features and want to be able to predict or generate similar structured objects given only their features. Finally, architectures combining discrete optimisation and learning fit within the neuro-symbolic framework. This framework gathers approaches capable of processing natural inputs – such as images, text or molecules – while simultaneously exhibiting extensive logical reasoning capabilities. Given the accumulating evidence that large language models are limited in their ability to perform logical reasoning (Bachmann and Nagarajan 2024; Lin et al. 2025; Ye et al. 2025; Zubic et al. 2025), neuro-symbolic systems are a promising alternative research direction in the quest for sparks-of-AGI (artificial general intelligence) (Ye et al. 2025).

In machine learning, the problem of predicting complex discrete objects (sequences, trees, graphs, ...) has been addressed in the field of Structured Output Prediction (SOP): given a training set of pairs $(\omega^\ell, \mathbf{y}^\ell)$, where $\omega^\ell$ is a suitable input data and $\mathbf{y}^\ell$ is a corresponding structured output, the aim is to predict a new high-quality structured output $\mathbf{y}$ from new input data $\omega$. In this setting, $\mathbf{y}$ can be predicted by arbitrary methods. In this paper, we are more specifically interested in architectures where $\mathbf{y}$ is an optimal solution of a predicted combinatorial optimisation (CO) problem instance with unknown objective function $F$ with parameters $\mathbf{c}$ and possibly unknown constraints $\mathscr{C}$. In a generative context, without retraining, the predicted CO instance can be arbitrarily constrained to produce optimal $\mathbf{y}$ that satisfy additional properties. Alternatively, additional objectives can be considered to produce optimal compromises.

This setting raises several challenges. First, the discrete nature of variables defines zero gradients almost everywhere, preventing direct learning by gradient-based optimisation. Second, predicting optimal solutions of CO problems using exact solvers or heuristic algorithms may be computationally demanding and automatic differentiation (autograd) from DL libraries is not applicable. The computational complexity is specifically problematic at training time, with massive training sets that may contain large NP-hard problem instances. We designed our approach to scale to such demanding situations for broad applicability. Our approach can also deal with partially observable settings, where observed solutions $\mathbf{y}^\ell$ may contain the values of only a subset of all variables.

Our contributions are threefold: we provide an extensive review of the literature, we introduce a neuro-symbolic architecture combining arbitrary deep learning layers with a final discrete NP-hard Graphical Model (GM) reasoning layer, and propose a new loss function that efficiently deals with logical information presented as constraints. We use discrete GMs (M. Cooper et al. 2020) as the reasoning language because GMs have been used to represent both Boolean functions or constraints (in propositional logic, constraint networks) and numerical functions or objectives (in partial weighted Max-SAT, cost function networks). For learning, a probabilistic interpretation of such models is available in stochastic GMs such as Markov Random Fields (MRFs), where infinite costs represent zero probabilities (infeasibility).

We exploit this probabilistic interpretation and target the optimisation of the negative loglikelihood (NLL) of the training set. Because of its intractability, we use an asymptotically consistent and scalable approximation of the NLL known as the Negative Pseudo-LogLikelihood (NPLL) (Besag 1975). We observe that the NPLL may become incapable of estimating many large costs (Montanari and Pereira 2009) and, therefore, of learning constraints. We analyze the reasons for this limitation and propose the E-PLL (Emmental Pseudo-LogLikelihood), which combines NPLL scalability with the ability to learn constraints. At inference, the neural architecture receives natural inputs and outputs a fully-parameterized discrete GM. This model can then be solved using any GM optimisation solver to produce a solution.

With this differentiable informative loss, our architecture efficiently learns constraints from natural inputs to, *e.g.*, solve the Visual Sudoku (Brouard et al. 2020; P.-W. Wang et al. 2019). Thanks to its ability to deal with unobserved variables, it can solve the associated symbol grounding problem without cheating (Chang et al. 2020). The use of an exact prover during inference provides 100% accuracy on hard symbolic Sudoku instances, even with small training sets. The same accuracy is obtained on a many-solution Sudoku data set (Nandwani et al. 2021) and more complex Futoshiki grids (Şen and Diner 2024). To show that our approach applies in a DFL context where constraints are known, we also tackle Min-Cut and Max-Cut instances where edge weights are represented as images. Finally, as a certificate of scalability, we train our architecture on a set of Protein Design instances with more than 1,000 variables with domain-size 20.

## 2 Related Works

*Inverse optimisation and constraint acquisition.* The idea of learning the parameters of optimisation problems from a set of solutions has been explored in various fields. Classical inverse optimisation (Chan et al. 2023) starts from a dataset of solutions $(\mathbf{y}^\ell)$, each being an exact optimum $\mathbf{y}^\ell \in \arg\min F(\mathbf{y}, \mathbf{c})$ with unknown parameters $\mathbf{c}$, that must be identified. Similarly, in constraint programming, constraint acquisition starts from sets of solutions, and possibly non-solutions, of an unknown constraint satisfaction (Beldiceanu and Simonis 2016; Bessiere et al. 2023) or satisfiability (M. Kumar et al. 2023) problem that needs to be identified. Compared to our setting, these approaches start from exact solutions/non solutions and cannot exploit arbitrary observable features $\omega$ such as Sudoku images. In data-driven inverse optimisation (Chan et al. 2023), the optimisation problem parameters $\mathbf{c}$ are split into observable and unobservable parameters $\mathbf{c} = \mathbf{c}_o \cup \mathbf{c}_u$. Starting from a dataset of pairs $(\mathbf{y}^\ell, \mathbf{c}_o^\ell)$, the aim is to predict $\mathbf{c}_u$ so that on average, empirically, each $\mathbf{y}^\ell$ is close, in distance or cost, to an optimal solution of $F(\mathbf{y}, \mathbf{c}_o^\ell \cup \mathbf{c}_o)$. This can be seen as a specific case of our setting, where the observed data $\omega$ is the exact observation of a fraction of the parameters instead of features known to be correlated with the complete problem definition (*i.e.,* both constraints and objective). However, existing approaches are dedicated to continuous (linear, convex, or conic) optimisation problems.

In its ability to learn constraints, our work is more closely related to the work of (Li, Y. Huang, et al. 2023) that combines neural networks with the MaxSAT Z3 solver (De Moura and Bjørner 2008): from the same input, they predict discrete constraints, by solving a continuous optimisation problem at training time. The first difference with our work lies in the target constraint language used. While we use arbitrary pairwise finite domain constraints, they use "unit" (with 0/1 coefficients) cardinality constraints on Boolean variables. These constraints can have arbitrary arity but take a very restricted form, which may require the introduction of additional latent variables (Li, Y. Huang, et al. 2023, § 6). The number of such constraints growing exponentially with the number of variables, a maximum number of learnable constraints, as well as the number of latent variables used, must be set a priori, adding to many hyper-parameters. A second difference lies in the theoretical scalability of both methods: we use an efficient convex loss while they rely on a "Difference of Convex" (DC) programming formulation. While defined on continuous variables, DC is NP-hard in general. Beyond the learning of logical constraints, our loss is also able to learn an objective function, which is required in the context of Decision-focused learning or Protein Design.

*Probabilistic Logic programming.* Deep learning has been combined with logic programming into several tools such as DeepProbLog (Manhaeve et al. 2018) or NeurASP (Yang et al. 2020), where deep learning is in charge of perception and Prolog/ASP is in charge of reasoning. These systems are mostly equipped with the ability of back-propagating the semantic loss (Xu et al. 2018), allowing for training neural nets using logical information, a task that differs from the task considered here.

### 2.1 Decision-Focused Learning

In many discrete optimisation under uncertainty problems, the target output $\mathbf{y}$ is the solution of a CO problem defined by a fixed set of constraints $\mathscr{C}$ and an objective function $F$ with unknown parameters $\mathbf{c}$. The parameters $\mathbf{c}$ can be predicted from historic data of features $\omega$, using a training set of pairs $(\omega^\ell, \mathbf{c}^\ell)$. In this setting, Predict-then-Optimize – or Prediction-focused-learning – methods (Mandi et al. 2024) learn to directly predict parameters $\hat{\mathbf{c}}$ from input features $\omega$ with a typical regression loss such as MSE, regardless of the downstream optimisation task. At inference, the fixed discrete model is solved with the predicted $\hat{\mathbf{c}}$. Instead, DFL approaches aim to predict parameters leading to "good" decisions that empirically optimize the average regret, the loss in objective caused by optimizing with $\hat{\mathbf{c}}$ instead of the true parameters $\mathbf{c}$.

DFL differs from our setting by the very nature of the training data. Still, since $\mathscr{C}$ is fixed, DFL training-set pairs $(\omega^\ell, \mathbf{c}^\ell)$ can be transformed into pairs $(\omega^\ell, \mathbf{y}^\ell)$ where $\mathbf{y}^\ell$ satisfies the constraints in $\mathscr{C}$ and optimizes $\mathbf{c}^\ell$, just by solving the optimisation problem defined by $\mathscr{C}$ and $\mathbf{c}^\ell$.

Thus, methods designed for SOP directly apply to DFL settings and have been *de facto* considered as DFL tools in recent reviews (Mandi et al. 2024). Existing methods have been classified in four families (Mandi et al. 2024): analytical differentiation of optimisation mappings, analytical smoothing of optimisation mappings, smoothing by random perturbations, and differentiation of surrogate loss functions. Differentiable optimisation layers have mostly been developed for convex or conic continuous optimisation problems. They can be applied to discrete optimisation problems after proper smoothing. This is, for instance, the case of SAT-Net (P.-W. Wang et al. 2019), which relaxes the discrete MaxSAT problem into a convex semi-definite programming problem. SAT-Net has been used to learn how to solve easy Sudoku and Visual Sudoku (from grid images). It has been pointed out that SAT-Net was ineffective if labels for observed digit images were not provided in the supervised training set (Chang et al. 2020).

*Smoothing with regularization.* Analytical smoothing is also needed for learning linear programming (LPs) problems because of the intrinsically discrete nature of the polytope's vertices. One standard smoothing is the addition of an $L_2$ regularization on solutions (Niculae and Martins 2020; Wilder et al. 2019). For integer LPs (ILPs), it is possible to relax them to LPs (Wilder et al. 2019), possibly with cutting plane strengthening (Ferber et al. 2020) for a better polytope approximation. Discrete (Boolean/One Hot Encoded) pairwise GMs, the formalism we use in this paper, are intrinsically quadratic in their discrete decision variables and linear in their parameters. They can be optimized by reduction to an 01LP, requiring the introduction of a quadratic number of variables. This 01LP model, relaxed to continuous variables, is known as the GM's "local polytope" (M. Cooper et al. 2020). Existing approaches for learning ILP could, therefore, be used to learn GMs. Due to their size, the associated (I)LP reductions are challenging to solve (Allouche et al. 2014; Hurley et al. 2016).

*Smoothing by random perturbation.* Besides smoothing by relaxation or regularization, *e.g.* in Fenchel-Young losses (Blondel et al. 2020), smoothing can be achieved by random perturbations (Berthet et al. 2020). These approaches rely on a probabilistic interpretation of the linear objective function using maximum-entropy exponential distributions. For discrete models, assuming minimisation, this usually means Gibbs/Boltzmann distributions with the probability $p(\mathbf{y}) \propto \exp(-\langle \mathbf{c}, \mathbf{y} \rangle)$. In arbitrary discrete settings, the resulting probability distributions suffer from the intractability of computing the normalizing constant $Z$. This can sometimes be simplified using the Gumbel trick (Huijben et al. 2022), which allows for estimating $Z$ using a series of calls to an optimisation oracle on a Gumbel-perturbed objective parameters $\mathbf{c}$ (Niepert et al. 2021) or using other perturbations (Berthet et al. 2020). These approaches can be interpreted as Fenchel-Young losses (Berthet et al. 2020; Blondel et al. 2020), where the perturbed solutions provide the regularization. The stochastic variants of GMs (M. Cooper et al. 2020; Koller and Friedman 2009) we rely on, Markov Random Fields, are decomposable log-linear models that rely on the same family of exponential distributions. Their log-likelihood is intractable for the same reason. It is known to be a Fenchel-Young loss (Blondel et al. 2020), often referred to as the CRF (Conditional Random Field) loss. We instead use and improve a tractable approximation of log-likelihood that exploits the decomposability of GMs.

*Differentiation of surrogate loss function.* These approaches try to optimize regret indirectly. The SPO+ loss (Elmachtoub and Grigas 2022) is an upper bound of regret with a gradient that resembles perturbed optimizers' gradients. A recent related approach, based on Noise Contrastive Estimation (Mulamba et al. 2021), exploits an incrementally built pool of feasible solutions in a probabilistic interpretation, as above.

## 2.2 Approaches without a Discrete Model

Other related approaches drift away from the DFL setting as they do not explicitly model or assume the existence of an underlying optimisation model that could be learned. A neural architecture directly learns to predict good solutions $\mathbf{y}$ from $\omega$. These approaches can naturally not guarantee constraint satisfaction and, unless properly conditioned, need to be retrained if constraints need to evolve. This area of research has recently developed with the observed incapacity of autoregressive Large Language Models (LLMs) to learn how to solve problems that require thinking ahead, such as Sudoku (Ye et al. 2025) or Zebra-like problems (Lin et al. 2025; Smith 1992; Zubic et al. 2025), as well simple path finding problems (Bachmann and Nagarajan 2024). The seminal Recurrent Relational Network paper (Palm et al. 2018) uses a Graph Neural Net-like network to learn how to solve Sudoku. Later, a recurrent Transformer architecture (Yang et al. 2023) was trained to solve Sudoku and Visual Sudoku. Recently, as an alternative to autoregressive LLMs, a discrete Denoising Diffusion Probabilistic Model (DDPM) was proposed to solve the same tasks (Ye et al. 2025), with a 100% accuracy and a related Hierarchical Reasoning model (G. Wang et al. 2025) shown to be able to solve extremely challenging Sudoku grids with 55% accuracy.

Stochastic GMs and deep learning have often been combined in machine learning and image processing in architectures that resemble ours, but with very different targets, such as semantic image segmentation (Z. Liu et al. 2017) or semi-supervised labeling (Qu et al. 2019).

Overall, in the context of learning how to solve NP-hard discrete optimisation problems, our approach is the only one that requires no solver call at training while learning a model that can be solved with guarantees, and the only one that targets an objective that is non-linear in its decision variables while being able to learn an objective together with constraints.

# 3 Preliminaries

## 3.1 Background on Discrete Graphical Models

A discrete graphical model (GM) is a concise description of a joint function of many discrete variables as the combination of many simple functions. Depending on the nature of the output of the functions (Boolean or numerical), and how they are combined and described, GMs cover a large spectrum of AI NP-hard reasoning and optimisation frameworks, including Constraint Networks, Propositional Logic as well as their numerical additive variants Cost Function Networks and partial weighted MaxSAT (M. Cooper et al. 2020; Schiex et al. 1995). Following cost exponentiation and normalization, these numerical joint functions can describe joint probability distributions, as done in Markov Random Fields (MRFs). In this paper, we use Cost Function Networks for their ability to express both numerical and logical functions. We assume here that cost functions take their value in $\bar{\mathbb{R}} = \mathbb{R} \cup \{\infty\}$.

*Notations.* In the rest of the document, we denote sequences, vectors, and tensors in bold. Variables are denoted in capitals with a given variable $Y_i \in \mathbf{Y}$ being the $i^{th}$ variable in the sequence $\mathbf{Y}$. An assignment of the variables in $\mathbf{Y}$ is denoted $\mathbf{y}$ and $y_i$ is the assignment of $Y_i$ in $\mathbf{y}$. $\mathbf{Y}_{-i}$ denotes the sequence of variables $\mathbf{Y}$ after removal of variable $Y_i$ and similarly for $\mathbf{y}_{-i}$ given $\mathbf{y}$. The domain of a variable $Y_i$ is a set denoted $D^i$ with $|D^i| \leq d$, the maximum domain size. For a sequence of variables $\mathbf{Y}$, we denote as $D^{\mathbf{Y}}$ the Cartesian product of all $D^i$ with $Y_i \in \mathbf{Y}$. A cost function over a subset of $\mathbf{Y}$ described by a tensor (cost matrix) over $\bar{\mathbb{R}}$ is called an *elementary* cost function.

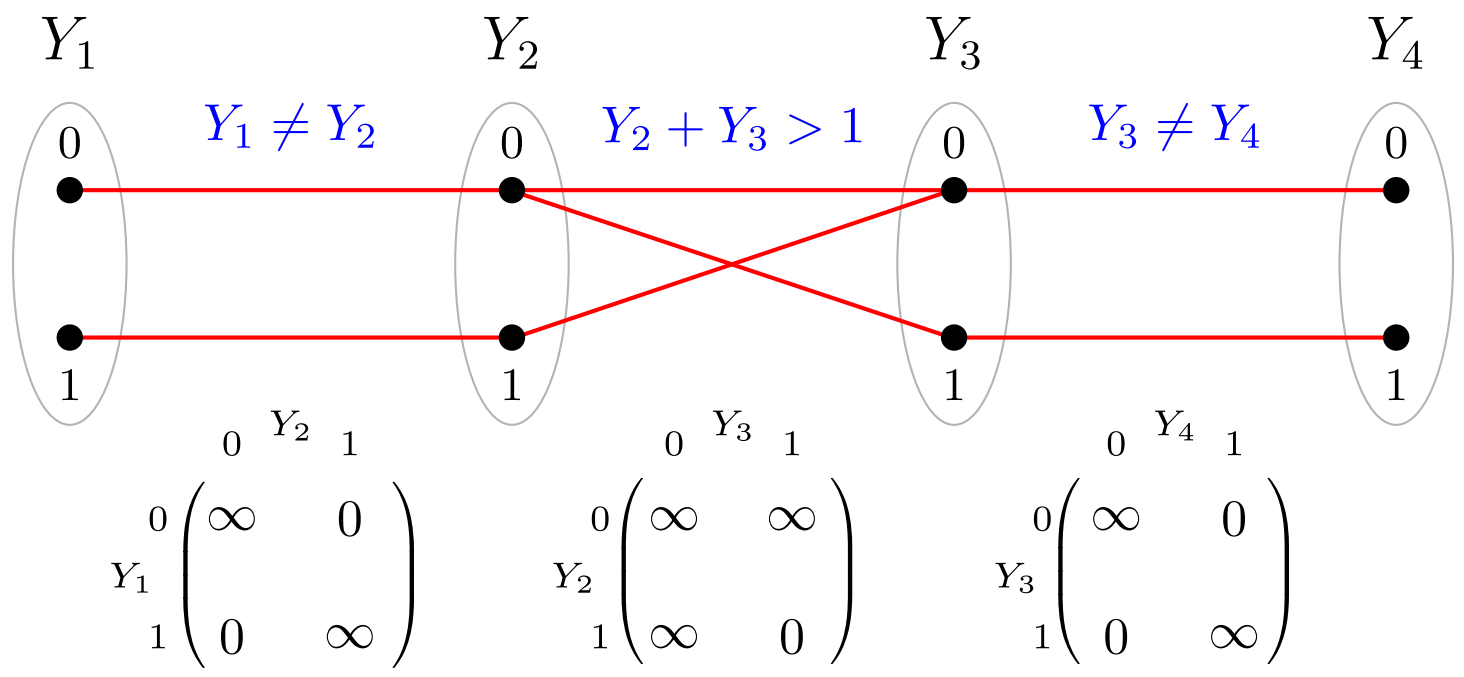

Fig. 1. Representation of Example 1. Variables are represented as ellipses, values as vertices. A red edge indicates an incompatibility, infinite cost, or zero probability for the corresponding pair of values. A constraint between two variables can be represented as a cost matrix, as illustrated below each constraint. TikZ code generated using Gemini Pro 2.5.

Definition 1. *Given a sequence* $\mathbf{Y} = \{Y_1, \dots, Y_n\}$ *of n finite domain variables, a cost function network* $\mathcal{M}$ *is defined as a set of elementary cost functions. It defines a joint cost function, also denoted* $\mathcal{M}(\cdot) = \sum_{F \in \mathcal{M}} F$*, involving all variables in* $\mathbf{Y}$*. The optimisation problem, known as the Weighted Constraint Satisfaction Problem (WCSP), is to find an assignment* $\mathbf{y}$ *that minimizes the joint function* $\mathcal{M}(\mathbf{y})$*. If* $\mathcal{M}(\mathbf{y}) < \infty$*,* $\mathbf{y}$ *is called a solution (a model in propositional logic).*

In stochastic GMs such as MRFs, the joint function $\mathcal{M}(\cdot)$ is used to define a joint probability distribution $P^{\mathcal{M}} \propto \exp(-\mathcal{M}(\cdot)) = \prod_{F \in \mathcal{M}} \exp(-F)$, requiring to compute a #P-hard normalizing constant.

A constraint is a cost function $F$ such that $F(\mathbf{y}) \in \{0, \infty\}$: it exactly forbids all assignments $\mathbf{y}$ such that $F(\mathbf{y}) = \infty$. When a given function $F$ is dominated by another function $F'$ (*i.e.*, $\forall \mathbf{y} \in \mathbf{Y}, F(\mathbf{y}) \leq F'(\mathbf{y})$, denoted $F \leq F'$), $F$ is known as a relaxation of $F'$. When $F \leq F'$ are constraints, we say that $F$ is a logical consequence of $F'$: whenever $F'$ is satisfied (equal to 0), $F$ is satisfied too. For a set of constraints $\mathbf{C}$, $F \in \mathbf{C}$ is redundant w.r.t. $\mathbf{C}$ iff $\mathbf{C}$ and $\mathbf{C} \setminus \{F\}$ define the same joint function. At a finer grain, we say $F$ is partially redundant if $\exists F' < F$ such that $(\mathbf{C} \setminus \{F\}) \cup \{F'\}$ and $\mathbf{C}$ define the same function.

Example 1. *Consider* $\mathbf{Y} = \{Y_1, Y_2, Y_3, Y_4\}$ *with domains* $\{0, 1\}$ *and* $\mathbf{C} = \{Y_1 \neq Y_2, Y_2 + Y_3 > 1, Y_3 \neq Y_4\}$ *represented in Figure 1. No constraint is redundant in* $\mathbf{C}$*, but in the context of the assignment* $\{Y_2 = 1, Y_3 = 1\}$*, the constraint* $Y_2 + Y_3 > 1$ *becomes redundant w.r.t.* $\mathbf{C}' = \mathbf{C} \cup \{Y_2 = 1, Y_3 = 1\}$ *. In the context of* $\{Y_1 = 0\}$*,* $Y_2 + Y_3 > 1$ *becomes partially redundant, as it could equivalently be replaced by the weaker* $Y_2 = Y_3$*. Due to redundancies, the observed values in a sample can create a context that makes some constraints redundant and, therefore, not learnable as its presence or absence does not change the problem semantics.*

For $n$ variables, a strictly pairwise graphical model $\mathcal{M}$ ($\forall F \in \mathcal{M}$, $F$ involves exactly 2 variables) can be described with $n(n-1)/2$ elementary cost functions with tensors (matrices) of size at most $d^2$. We denote by $\mathcal{M}[i, j]$ the tensor describing the cost function between variables $Y_i$ and $Y_j$ in $\mathcal{M}$. In practice, when $n$ is large, we can model a restricted number of cost functions (see Subsection 5.4 on protein design).

## 3.2 Problem Statement

In this work, we assume that we observe samples $(\omega, \mathbf{y})$ of the values $\mathbf{y}$ of the variables $\mathbf{Y}$ as low-cost solutions of an underlying constrained optimisation problem with parameters influenced by natural inputs $\omega$. From a data set $\mathbf{S}$ of pairs $(\omega, \mathbf{y})$, we want to learn a model $N$ (in our case, a neural network) that predicts a pairwise graphical model $\mathscr{M} = N(\omega)$ such that $\mathbf{y} \in \arg\min_{\mathbf{y}\in D^{\mathbf{Y}}} N(\omega)(\mathbf{y})$. This graphical model $\mathscr{M} = N(\omega)$ defines the last layer of our hybrid neural+graphical model architecture (see Figure 2).

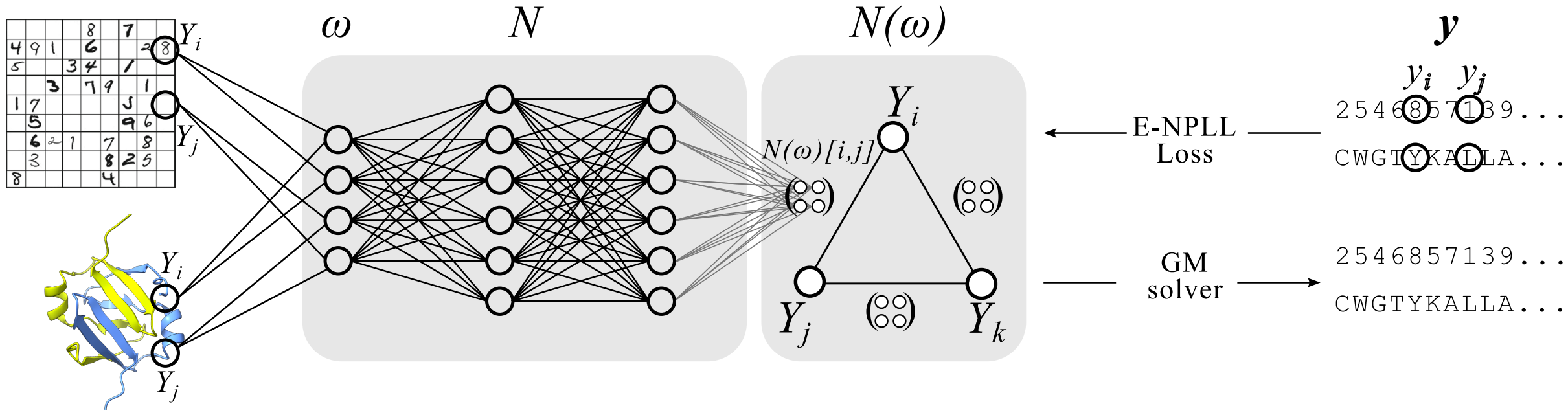

Fig. 2. Our hybrid learning architecture: natural inputs $\omega$ (left) feed a neural net $N$ in charge of predicting all pairwise cost functions $F_{ij}$ of the GM $\mathscr{M} = N(\omega)$. To learn $N$, we back-propagate solutions $\mathbf{y}$ for $(\omega, \mathbf{y}) \in \mathbf{S}$ through the E-PLL loss function (no solver is used). At inference, $N(\omega)$ can be directly fed to any GM solver, be it exact, based on a scalable relaxation or a (meta)-heuristic. This is illustrated here on 2 possible problems: a visual Sudoku problem (top) and a protein design problem (bottom).

In terms of supervision, all variables in $\mathbf{Y}$ are usually observed, but we will allow for partial observation in Subsection 4.1. We also want to exploit any information that would be available on elements of $\omega$. Some of these natural inputs may be direct constraints or assignments of variables in $\mathbf{Y}$ that can be directly incorporated into the GM $N(\omega)$, others may be known to influence only a subset of all variables $\mathbf{Y}$. In the symbolic Sudoku problem, a partially filled grid of numbers is observed in $\omega$. Each observed value in the grid is known to determine the value of its corresponding variable.

Assuming the data set $\mathbf{S}$ contains i.i.d. samples from an unknown probability distribution $P(\mathbf{Y}|\omega)$, a natural loss function for the GM $N(\omega)$ is the asymptotically-consistent negative logarithm of the probability $\mathrm{NLL}(\mathbf{S}) = -\log(\prod_{(\omega,\mathbf{y})\in\mathbf{S}} P^{N(\omega)}(\mathbf{Y} = \mathbf{y}))$ of the observed samples. This negative loglikelihood is, however, intractable because of the #P-hard normalizing constant needed to compute the probability. We instead rely on the tractable negative pseudo-loglikelihood (Besag 1975) $\mathrm{NPLL}(S) = \sum_{(\omega,y)\in S} -\log(\prod_i P^{N(\omega)}(Y_i = y_i|\mathbf{y}_{-i}))$. The NPLL works at the level of each variable $Y_i$, in the context of $\mathbf{y}_{-i}$, the assignment of all other variables. It requires only normalization over one variable $Y_i$, a computationally easy task. Generalized variants of the NPLL use subgroups of variables (F. Huang and Ogata 2002) instead of single variables, at the cost of additional complexity. In Appendix B.1, we give a simple proof of the asymptotic consistency of the basic PLL (a folklore result), under a strict convexity assumption, while Appendix B.2 reframes the PLL as a Fenchel-Young loss.

Property 1. *Due to the decomposability of the joint function $\mathscr{M}(\cdot)$, we have* $P^{\mathscr{M}}(Y_i|\mathbf{y}_{-i}) = \mathrm{softmax}(-m_i(Y_i))$ *where* $m_i(Y_i) = \sum_{j\neq i} \mathscr{M}[i,j](Y_i, y_j)$. *In a message passing interpretation,* $m_i(\cdot) \in \bar{\mathbb{R}}^{|D^i|}$ *represents the sum of all messages received from neighbor variables $Y_j$ through the incident functions $\mathscr{M}[i,j]$, given $Y_j = y_j$. Computing the NPLL is in $O(n(n-1)d)$ per sample and epoch. It can easily be vectorized (computed independently on each variable).*

The NPLL enables scalable training from natural inputs in the context of an underlying NP-hard GM optimisation problem. However, the proofs of asymptotic consistency of the NPLL (Besag 1975; Geman and Graffigne 1986) rely on identifiability or strict convexity assumptions (see Appendix B.1) that do not hold in the context of constraints (zero probabilities) as redundant constraints can be added or not without influencing the joint function $\mathscr{M}(\cdot)$. Unsurprisingly, the NPLL is known to perform poorly in the presence of large costs (F. Huang and Ogata 2002;

Montanari and Pereira 2009). Empirically, we observed that the resulting architecture completely fails at solving even the simplest symbolic Sudoku problem where $\omega$ contains the digits from the unsolved Sudoku grid and $\mathbf{y}$ is the corresponding solution.

## 4 The E-PLL

To understand the incapacity of the NPLL to deal with large costs, it is interesting to look at the contribution of every pair $(\omega, \mathbf{y})$ to the gradient $\frac{\partial NPLL}{\partial N(\omega)[i,j](v_i,v_j)}$ of the NPLL for a given pair of values $(v_i, v_j)$ of a pair of variables $(Y_i, Y_j)$.

Property 2 (see Appendix A). *The contribution of a pair $(\omega, \mathbf{y})$ to the gradient $\frac{\partial NPLL}{\partial N(\omega)[i,j](v_i,v_j)}$ is*

$$[\mathbb{1}(y_i = v_i, y_j = v_j) - P^{N(\omega)}(v_i|\mathbf{y}_{-i})\mathbb{1}(y_j = v_j)] + [\mathbb{1}(y_i = v_i, y_j = v_j) - P^{N(\omega)}(v_j|\mathbf{y}_{-j})\mathbb{1}(y_i = v_i)]$$

where $\mathbb{1}$ is the indicator function. The two terms in the gradient above come from NPLL terms computed on variables $Y_i$ and $Y_j$, respectively.

Example 2. *Consider our previous example with four variables, $C = \{Y_1 \neq Y_2, Y_2 + Y_3 > 1, Y_3 \neq Y_4\}$ and $\mathbf{y} = (0, 1, 1, 0)$. We focus on the variables $Y_{i=2}$ and $Y_{j=3}$ and assume that $C$ should hold under $\omega$, which means that the pair $(Y_2 = 0, Y_3 = 0)$ should be predicted as forbidden. With $\mathbf{y} = (0, 1, 1, 0)$, we have, $\mathbb{1}(y_2 = 0, y_3 = 0) = 0$.*

*Assume now that ongoing learning is already predicting a high cost for the forbidden pairs $(Y_1 = 0, Y_2 = 0)$ and $(Y_3 = 0, Y_4 = 0)$, then both $P^{N(\omega)}(Y_2 = 0|\mathbf{y}_{-2})$ and $P^{N(\omega)}(Y_3 = 0|\mathbf{y}_{-3})$ will be close to zero and the gradient will be close to zero too. This will lead to a negligible (if any) change in the cost of the pair $(Y_2 = 0, Y_3 = 0)$: learning will be blocked or tremendously slowed down: the fact that, in the context of $(\omega, \mathbf{y})$, the forbidden pair $(Y_2 = 0, Y_3 = 0)$ is redundant w.r.t. already identified forbidden pairs $(Y_1 = 0, Y_2 = 0)$ and $(Y_3 = 0, Y_4 = 0)$ effectively prevents learning to predict the cost $N(\omega)[2, 3](0, 0)$.*

The issue with the NPLL lies in the dynamic of the stochastic gradient optimisation: the early identification of some high costs under $\omega$ will prevent the increase of other significant costs which are redundant in the context of the observed $\mathbf{y}$, but not redundant in the global unconditioned problem.

Inspired by "dropout" in deep learning (Srivastava et al. 2014), we introduce the Emmental NPLL (E-PLL) as an alternative to the NPLL that can learn all constraints (infeasibilities) present in $S$.

Definition 2. *Given a GM $\mathcal{M}$ over variables $\mathbf{Y}$ and $\mathbf{H} \subset \mathbf{Y}$, we denote by $\mathcal{M} - \mathbf{H}$ the graphical model derived from $\mathcal{M}$ by replacing all cost functions involving a variable in $\mathbf{H}$ by a constant $0$ function. The E-PLL can then be defined as*

$$\text{E-PLL}(\mathbf{y}) = -\sum_{Y_i \in \mathbf{Y}} \log(P^{(N(\omega)-\mathbf{H}_i)}(Y_i = y_i|\mathbf{y}_{-i}))$$

*where each $\mathbf{H}_i$ is a random subset of $\{1, \ldots, n\} \setminus \{i\}$.*

The idea of the E-PLL follows directly from the previous gradient analysis: to prevent a combination of incident functions with already-learned high cost from shrinking gradients, we mute a random fraction of the incident functions (or equivalently, variables, as we learn pairwise functions). It can be understood intuitively on the Sudoku example: the NPLL asks to predict one cell given the rest of the grid. Once all the row constraints are learnt, one missing cell will always be predicted correctly and the loss is zero. Yet for inference all the constraints are needed. To force the neural network to also learn column and square constraints, the E-PLL randomly masked some of the cells.

The E-PLL is used alongside an L1 regularization on the output of the learned network $N(\omega)$ to favour sparsity. This also makes the GM optimisation problem easier to solve. The complete training loss is then:

$$\mathscr{L}(\mathbf{y}) = \text{E-PLL}(\mathbf{y}) + \lambda \cdot ||N(\omega)||_1$$

Because the E-PLL is designed to fight the side effects of redundant constraints on gradients, we expect it to learn a GM $N(\omega)$ with all redundant pairwise constraints.

### 4.1 Dealing with Unobserved Variables

The E-PLL is defined only over complete assignments. In the context of missing data, in every sample $(\omega, \mathbf{y})$, $\mathbf{y}$ may be a partial assignment. For example, in the visual Sudoku, each image of a digit in the grid is known to represent the value of a single variable. If the values of initial hints are observed in $\mathbf{y}$, they provide direct supervision and thus grounding information (Chang et al. 2020) when training digit recognition. Simultaneous learning of Sudoku rules and digit recognition requires to mask initial hints in the sequence, leading to unobserved variables. We tackle this Visual Sudoku problem in Subsection 5.2.

A usual strategy to deal with missing data is to rely on variants of the expectation-maximization (EM) algorithm (Qu et al. 2019), where expectations of sufficient statistics (Kholevo 2001) are used in the likelihood maximisation algorithm. In our case, this would require computing expensive #P-hard marginals. We instead rely on a simpler, NP-hard, imputation procedure where the values of the missing variables are obtained by optimizing the joint function defined by $N(\omega)$, all variables observed in $\mathbf{y}$ being assigned to their values in $\mathbf{y}$. The resulting complete assignment is then used instead of the partial $\mathbf{y}$. During training, this strategy requires one NP-hard oracle call per sample with missing data: training scalability is, in principle, lost. This brings our approach closer to the Nesy-programming (Li, Y. Huang, et al. 2023) approach, which allows for latent variables and needs to solve a (generally NP-hard) “Difference of Convex” problem.

When the fraction of unobserved variables remains limited, the solved problems are simple, with just a few unassigned variables. In our experiments, we use an exact GM solver to obtain a single imputation. For increased scalability, heuristic or approximate solvers (Durante et al. 2022; Mladenović and Hansen 1997) could be used instead, with multiple imputations (Little and Rubin 2019) if necessary.

## 5 Experiments

We demonstrate the versatility of our architecture by testing it on a variety of learning tasks:

(1) Learning the constraints of purely-symbolic logical puzzles: Sudoku and Futoshiki, with one or many solutions (Nandwani et al. 2021)
(2) Learning the constraints of logical puzzles (Sudoku) with visual inputs, a task known as Visual Sudoku (P.-W. Wang et al. 2019). We consider the ungrounded setting where the observed digit images in $\omega$ are missing in the corresponding $\mathbf{y}$.
(3) Learning the objective of contextual Min-Cut and Max-Cut problems. In this DFL-like setting, we observe that the E-PLL implicitly minimizes regret.
(4) Simultaneously learning the constraints and the objective of a real-world problem: designing new proteins.

Reported training times have been measured using a Nvidia RTX-2070 Super GPU with 8GB of VRAM and a 4.2 GHz AMD Ryzen 9 5900X CPU with 32 GB of RAM. During our measures, we noticed that the training accuracies were often slightly below those reported in the original papers. For all those that could run on our hardware with training times below 50 hours, we reran them twice to collect a total of 3 accuracies (including the paper result). For these, we report the min/max accuracies observed (when they differ). Our code is run with PyTorch 2.6 and PyToulbar2 0.0.0.4. We use the Adam optimizer with a weight decay of $10^{-4}$ and a learning rate of $10^{-3}$ (other parameters take default values). An L1 regularization with multiplier $\lambda = 2.10^{-4}$ is applied on the cost matrices $N(\omega)[i, j]$. Code and data are available at https://github.com/mdefresne/emmental_pll/.

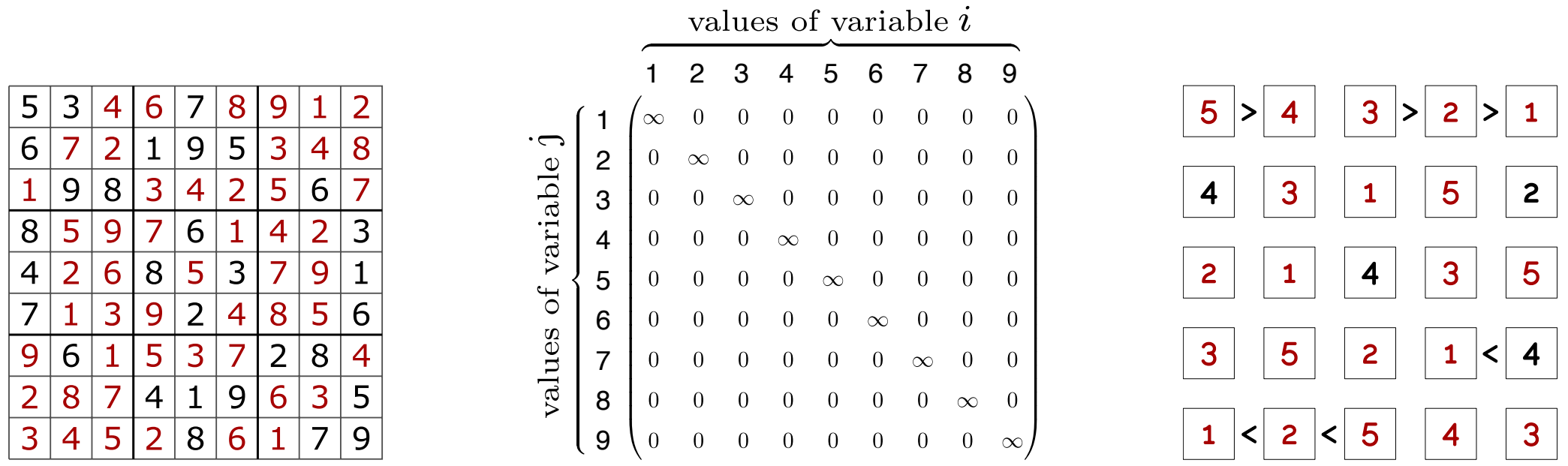

Fig. 3. A solved Sudoku grid (left), the cost function representing one Sudoku rule (middle), and a solved Futoshiki grid (right)

### 5.1 Learning how to Play Logic Puzzles

The NP-complete Sudoku problem is a classical logical reasoning problem that has been repeatedly used as a benchmark in a "learning to reason" context (Amos and Kolter 2017; Bessiere et al. 2023; Brouard et al. 2020; Defresne et al. 2023; Li, Y. Huang, et al. 2023; Palm et al. 2018; P.-W. Wang et al. 2019; Yang et al. 2023; Ye et al. 2025). The task is to learn how to solve new Sudoku grids from a set of solved grids, without knowing the game rules. On this task and two variants, we show that the E-PLL enables to learn the exact rules of the puzzles with high time-and-data-efficiency.

#### *5.1.1 Experimental Setting.*

*Sudoku.* Given samples $(\omega^\ell, \mathbf{y}^\ell)$ of initial and solved Sudoku grids, we want to learn how to solve new grids. Sudoku players know that Sudoku grids can be more or less challenging. As one could expect, it is also harder to train how to solve hard grids than easy grids (Brouard et al. 2020). We use the number of initially filled cells (hints) as a proxy for the problem's hardness, a grid with few hints being hard. The minimal number of hints required to define a single solution is 17, defining the hardest single-solution Sudoku grids. We use the RRN data set (Palm et al. 2018), composed of single-solution grids with 17 to 34 hints. We use 100 grids for training, and 32 for validation (all hardness). As in (Palm et al. 2018), we test on the hardest 17-hints instances, 1,000 in total.

Recently, a new *Sudoku-Extreme* dataset (G. Wang et al. 2025) was assembled based on the number of backtracks needed by the *tdoku* solver (github.com/t-dillon/tdoku) to solve instances. In (G. Wang et al. 2025), 1,000 extreme grids are used and augmented 1,000 times by a dedicated augmentation procedure, reaching a 1,001,000 grids training set. We use 10 extreme grids from the same data set, augmented 99 times only, with same augmentation code, to reach a final training set of 1,000 grids for training. For validation, we use the same validation subset as (G. Wang et al. 2025).

*One-of-Many solutions.* Published Sudoku grids have only one solution. We also consider a Sudoku benchmark used in (Nandwani et al. 2021), where each grid has more than one solution. For each grid, the set of solutions is only partially accessible during training: at most 5 of them are present in the training set. The aim is to be able to predict any of the feasible solutions (all of them are known for testing). We use 100, 32, and 256 grids of the data set from (Nandwani et al. 2021) respectively for training, validating, and testing. During training, one of the 5 available solutions for a given grid is randomly selected.

Table 1. Average performance over 10 initializations.

| Approach | Dataset | Training time (s) | Runs with 100% test grids solved |
|---|---|---|---|
| NPLL | Sudoku (unique) | - | 0% |
| E-PLL | Sudoku (unique) | $215 \pm 60$ | 100% |
| E-PLL | Sudoku (many) | $250 \pm 46$ | 100% |
| E-PLL | Futoshiki | $1839 \pm 7$ | 90% |

*Neural architecture.* A $9\times 9$ Sudoku grid is represented as 81 cell coordinates with a possible hint when available. Each cell is represented by a GM variable with domain $\{1, \ldots, 9\}$. For $N$, we reuse the same architecture (Defresne et al. 2023), but we drastically reduce the number of parameters. We use a Multi-Layer Perceptron (MLP) with 4 hidden layers of 64 neurons and residual connections (He et al. 2016) every 2 layers. It receives the pairs of coordinates of pairs of cells $(Y_i, Y_j)$ and predicts all pairwise cost matrices $N(\omega)[i, j]$. Hints are used to set the values of their corresponding variable in $N(\omega)$. Performances are measured by the percentage of correctly filled grids; no partial credit is given for individual digits.

*Futoshiki.* is another grid-based puzzle. As for Sudoku, no digit can be repeated twice in a row or column. In addition, the solution must respect inequalities between some pairs of cells (see Figure 3). We generated our own dataset, composed of $5 \times 5$ grids (for details, see annex C.1). We used $1,000$ training grids, 64 for validation and 200 for testing. The grid representation is similar to Sudoku: each pair of cells is represented by its coordinates (row and column indices) and an additional feature representing the inequality (1 if $i > j$, $-1$ is $i < j$ and 0 if no inequality). The same neural net as for Sudoku is used.

*Expected rules.* The neural architecture predicts all rules through the costs matrices $N(\omega)[i, j]$. For pairs of cells on the same row, column or 3 sub-square, we expect soft difference-like cost function to be predicted (a matrix with a strictly positive diagonal and zeros) that prevents the use of the same value for the two variables. It is illustrated in Figure 3. With the L1 regularization on the output of the neural net, other cost matrices should be 0, indicating the absence of a pairwise constraint. For Futoshiki, cost matrices are less sparse than for Sudoku. The diagonal still contains positive costs to prevent 2 identical digits, but it also contains positive costs on the lower (resp. upper) triangular part in case of superior (resp. inferior) inequality constraint.

*5.1.2 Test Results.* We first train our network with the regular NPLL loss. As expected, it learns only a subset of the rules that suffices to make all other rules redundant: for a cell $Y_i$ in the context of $\mathbf{y}_{-i}$, a single clique of difference constraints for every row (or column, or square) is sufficient to determine the value of the cell $Y_i$ from $\mathbf{y}_{-i}$, creating vanishing gradients for all other constraints that are instead estimated as constant 0 matrices. On the test set, inference completely fails.

We replaced the NPLL by the E-PLL, ignoring messages from $k$ randomly chosen other variables. In terms of accuracy, the training is largely insensitive to the value of the hyperparameter $k$ (see Table 6 in the Appendix) as long as it is neither 0 (regular NPLL) nor close to $n-1$ (no information). However, larger values of $k$ tend to lead to longer training. We set $k = 10$ for all Sudoku experiments. In this case, training takes only 2 minutes on a single CPU. At inference, the predicted $N(\omega)$ leads to 100% accuracy on all the hard test-set grids.

Since the E-PLL never compares a solver-produced solution to the provided solution $\mathbf{y}$, it is not sensitive to the existence of many solutions. Therefore, training on the dataset with multiple solutions yields similar results: one of the feasible solutions is predicted for 100% of the test grids. We come to a similar observation with Futoshiki.

Table 2. Accuracies of related works. They are sorted by type of approach: pure Deep Learning (DL), reasoning (CO for combinatorial optimisation), and relaxation of reasoning (Relax), which can be combined with ML or DL. The '# hints' gives the average hardness of the test set. Param. is the number of parameters of the neural network. HRM train-time is extracted from (G. Wang et al. 2025) running on a fast RTX 4070M as their code requires a recent GPU.

| Type | Approach | Acc. | #hints | Train set | Param. | Train time (h) |
|---|---|---|---|---|---|---|
| DL | RRN (Palm et al. 2018) | 96.6% | 17 | 180,000 | 200k | $> 50$ |
| | Rec. Trans. (Yang et al. 2023) | 96.7% | 17 | 180,000 | 211k | $> 50$ |
| | Rec. Trans. | 76.2–78.2% | 17 | 9,000 | - | 1.8 |
| | DDPM (Ye et al. 2025) | 99.2–100% | 33.8 | 100,000 | 6M | 13.6 |
| | DDPM | 0.2% | 17 | - | - | - |
| | HRM (G. Wang et al. 2025) | 55% | 24.8 | $1,000 \times 1,001$ | 27M | >10h |
| Relax+DL | SATNet (P.-W. Wang et al. 2019) | 95.1–99.8% | 36.2 | 9,000 | 600k | 2.9 |
| | SATNet | 86.1–86.2% | 17 | - | - | - |
| CO | (Bessiere et al. 2023) | **100%** | - | 200 | - | 0.01 |
| CO + ML | (Brouard et al. 2020) | **100%** | 17 | 9,000 | - | 1.5 |
| CO+DL | Hinge (Defresne et al. 2023) | **100%** | 17 | 1,000 | 180k | >50 |
| | **E-PLL** (ours) | **100%** | 17 | **100** | 22k | 0.05 |
| | **E-PLL** (HRM dataset) | **100%** | 24.8 | **10** × 100 | 22k | 0.04 |

For inference, a threshold is applied on all the cost functions: all costs below 1 are set to 0. All of the test grids were correctly solved.

In Table 2, we compare our results with related approaches that learn how to solve Sudoku. Pure Deep Learning methods, Recurrent Relational Network (RRN) (Palm et al. 2018), Recurrent Transformer (Yang et al. 2023), a Hierarchical variant (G. Wang et al. 2025) and Denoising Diffusion Probabilistic Models (DDPMs) (Ye et al. 2025), require orders of magnitude more data and fail to solve some of the hardest puzzles. The accuracy of Recurrent Transformers drops by roughly 20% when the amount of data is reduced to 9,000 samples. While DDPMs (Ye et al. 2025) solve Sudokus (from an easy Kaggle dataset) quite reliably, they completely fail on hard Sudokus. Adding a convex relaxation of Max-SAT reasoning optimisation layer, SATNet (P.-W. Wang et al. 2019) becomes much more data-efficient. Still, it fails to solve some of the easy grids, and its accuracy drops significantly on hard grids, below that of pure DL approaches.

Using a discrete solver alone (Bessiere et al. 2023), or combined with ML (Brouard et al. 2020), it is possible to solve all test instances reliably from properly learned rules, offering far more efficient training than DL-based approaches. This also extends to the Sudoku-extreme dataset, where only 10 grids with 99× augmentation suffice to reach a 100% accuracy, compared to 1, 000 grids with 1, 000× augmentation needed to reach 55% accuracy by the Hierachical Reasoning Model (G. Wang et al. 2025).

However, these non-end-to-end differentiable approaches cannot directly exploit natural inputs, as in the Visual Sudoku or protein design tasks described below. Hybrid approaches combine both DL and exact reasoning. A follow-up work (Li, Guo, et al. 2023) of SATNet, extracts explicit logical rules from SATNet, enabling reliable solving on $4 \times 4$ grids. On $9 \times 9$ grids, it learns hundreds of thousands of clauses, leading to unsolvable instances. Approaches that embed an exact solver during training, as the structured Perceptron/Hinge losses, can solve any grid (Defresne et al. 2023), at the cost of an excruciatingly long training time. Overall, our approach stands out for its perfect reliability, its data-efficiency, and its low training time.

*5.1.3 Retrieving Exact Constraints.* After learning, logical constraints can be retrieved with two alternative strategies. When constraints only need to be learned, a threshold can be applied (as done for Futoshiki): learnt costs below the threshold are set to 0 and costs above are set to $\infty$ (hard constraint). The second strategy is cost function hardening (Brouard et al. 2020), which has the advantage of requiring no parameter and also of preserving the learned objective (if needed). Non-zero learned costs are considered in decreasing order and set to $\infty$ if the corresponding value combination is not observed in any of the training set. This is repeated until a contradiction is found.

When applying cost function hardening to the output of the neural network trained for the Sudoku task, all of the 810 pairwise constraints are predicted, with no additional constraints. Therefore, the exact rules are learnt and we can be confident that the accuracy of 100% observed on the test set extends to any Sudoku instance. In the many-solution setting, hardening enables a complete enumeration of all feasible solutions for any test instance (instead of finding just one maximum probability assignment). From a learning point of view, predicting a discrete model improves interpretability, as one can look at the learnt constraints to understand the decision.

## 5.2 Visual Sudoku

*Task.* Our previous examples show the benefits of exploiting inputs $\omega$. To explore this capacity more deeply, we tackle the Visual Sudoku problem in which hints are MNIST images. The goal is to simultaneously learn how to recognize digits and how to play Sudoku. Therefore, we add an untrained convolutional neural network (CNN) to our previous architecture in order to recognize hand-written hints. The logits predicted by the CNN are negated and interpreted as a unary cost function on each variable $Y_i$ with a hint. The GM $N(\omega)$ produced comprises the pairwise cost functions predicted by the ResMLP (as before) and the unary cost functions. This GM is fed to our regularized E-PLL loss for back-propagating solutions.

*Data.* Our data set is obtained from the symbolic Sudoku data set by replacing hints with arbitrarily selected MNIST images, as in (Brouard et al. 2020; Topan et al. 2021). We use grids from the RRN dataset (9,000 for training, 64 for validation), as they are much more challenging than the SATNet dataset (average 36.2 hints) (P.-W. Wang et al. 2019). The test set contains 100 grids of each difficulty (from 17 to 34 initial hints).

*Grounded and ungrounded Visual Sudoku.* The original Visual Sudoku task (P.-W. Wang et al. 2019) presented a subtle form of data leakage (Chang et al. 2020): each hint digit image in the input $\omega$ had a corresponding label in the output $\mathbf{y}$, and it was possible to ground every image into its specific meaning directly. In the ungrounded Visual Sudoku task, the training set contains no information on the labels of the cells with hints' images. We handle this as missing data, using imputation to predict complete assignments from incomplete assignments $\mathbf{y}$ in the training set. During the first epochs, the predicted rules are mostly random, making the imputation longer. Therefore, we first restrict training to grids with few initial hints: only 17 on first epoch, then less than 20 on the second, 30 on the third and so on until all training grids are included. Moreover, we decay the learning rate of both neural nets by dividing it by 10 at epochs 6 and 8, training for 20 epochs.

*Test.* After visual Sudoku training, we assessed the accuracy of the CNN alone on MNIST digits, the percentage of correctly-filled grids when no correction of MNIST classification is allowed, and the final percentage of correct grids with corrections by the solver. In Table 3, we only compare methods tackling the grounded problem, where no label is available for hints. The resulting perception architecture is not as accurate as competitors, with a 1% lower accuracy, leading to far fewer properly filled grids. Yet, the Sudoku rules are learnt properly, enabling the solver to correct digit missclassification inconsistent with the rules. *In fine*, 20% of the grids are corrected, and the E-PLL reaches a competitive 93.3% of visual grids solved, with a training time reduced by 32%.

Table 3. Grounded Visual Sudoku performance (balanced RNN test set). Percep. refers to the percentage of grids correctly filled without modifying the recognized digits, while Solved is the percentage of correct grids after correction of misclassified digits.

| Approach | MNIST accuracy | Percep. | Solved | Training ($h$) |
|---|---|---|---|---|
| Rec. Trans (Yang et al. 2023) | 99.4% | 74.8% | 59.0–75.6% | 5.1 |
| NeSy. Prog. (Li, Y. Huang, et al. 2023) | 99.6–99.7% | 90.7–93.1% | 92.2–94.4% | 4.7 |
| **E-PLL** (Ours) | 98.8% | 69.3–72.9% | 92.1–93.4% | 3.2 |

### 5.3 Application in a DFL Setting with Min-Cut and Max-Cut

Although our method was originally designed to learn on SOP datasets $(\omega^\ell, \mathbf{y}^\ell)$, with no intermediate supervision, it can also be applied to DFL tasks with known constraints. To illustrate the versatility of our approach, we designed a new DFL task based on the Min-Cut and Max-Cut problems. A graph $G = (V, E)$ is given, where each edge $e \in E$ is associated with a capacity. To include natural inputs, each capacity is represented by an image of a bridge, either stone, wood or rope„ with decreasing respective capacities of 5, 2, and 1. The goal is to find a cut of the graph with minimum or maximum capacity.

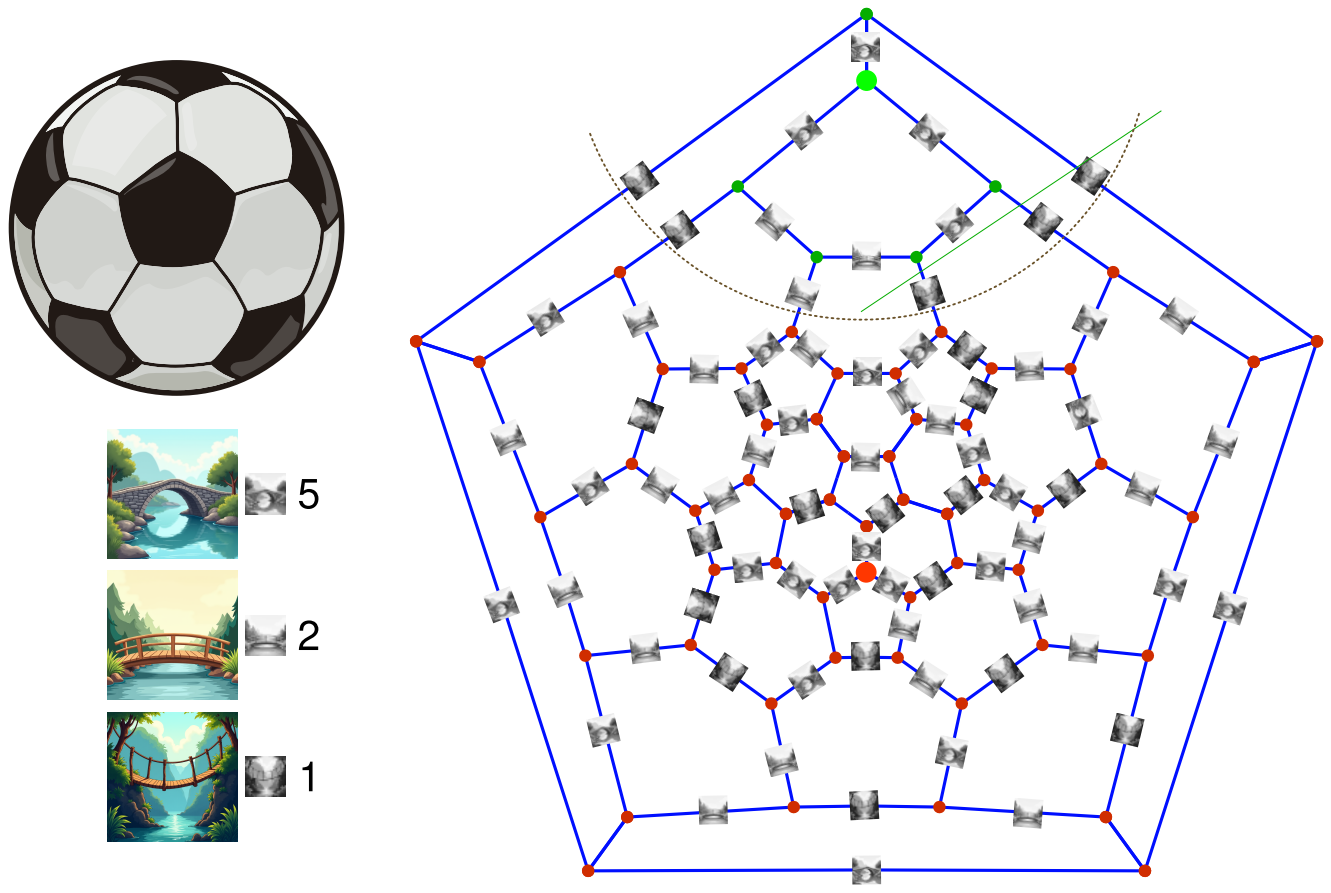

Fig. 4. The truncated icosahedron as a solid (top left) and as a planar graph (right). The randomly selected source and opposite sink vertices are shown in green and red, respectively. Each edge is associated with one of the 3 bridge images, reduced to $28 \times 28$ grayscale pixels. A minimum cut is identified by the dotted gray arc with green/red vertices identifying the two sides of the cut, here of capacity $7 = 1 + 1 + 2 + 1 + 1 + 1$. Bridge images by Flux 1.1-dev, planar graph from Wikipedia, solid image by Freepik.

To build training samples, we use the graph of a truncated icosahedron, an Archimedean solid with a planar graph, and select a source vertex randomly. The sink node is set to the opposite vertex in the solid. To avoid trivial cuts, edges incident to the source and sink nodes are associated with stone-bridge images. The remaining edges are assigned a random image/capacity (stone, wood, or rope). All images are noised before normalization with a Gaussian noise of mean 0 and standard deviation of 10 (for 256 levels, 8 bit images).

The Min/Max cut problems can be easily encoded in a GM where a Boolean variable is associated with every vertex, indicating its side of the cut. For every edge $e$ in $E$, a cost function proportional to a $2 \times 2$ identity matrix $I_2$ (Max-Cut) or $(1 - I_2)$ matrix (Min-Cut) is used. The multiplicative factor represents the cost to pay, if the edge is cut (Min-Cut) or not (Max-Cut). The DFL dataset $(\omega^\ell, \mathbf{c}^\ell)$ is turned into a SOP dataset $(\omega^\ell, y^\ell)$ by solving each instance $(\omega^\ell, \mathbf{c}^\ell)$ with the exact GM solver toulbar2 (Hurley et al. 2016). The $(1 - I_2)$ matrix being submodular, Min-Cut, the dual of Max-flow, is solved efficiently (M. C. Cooper et al. 2010). Max-Cut is NP-hard.

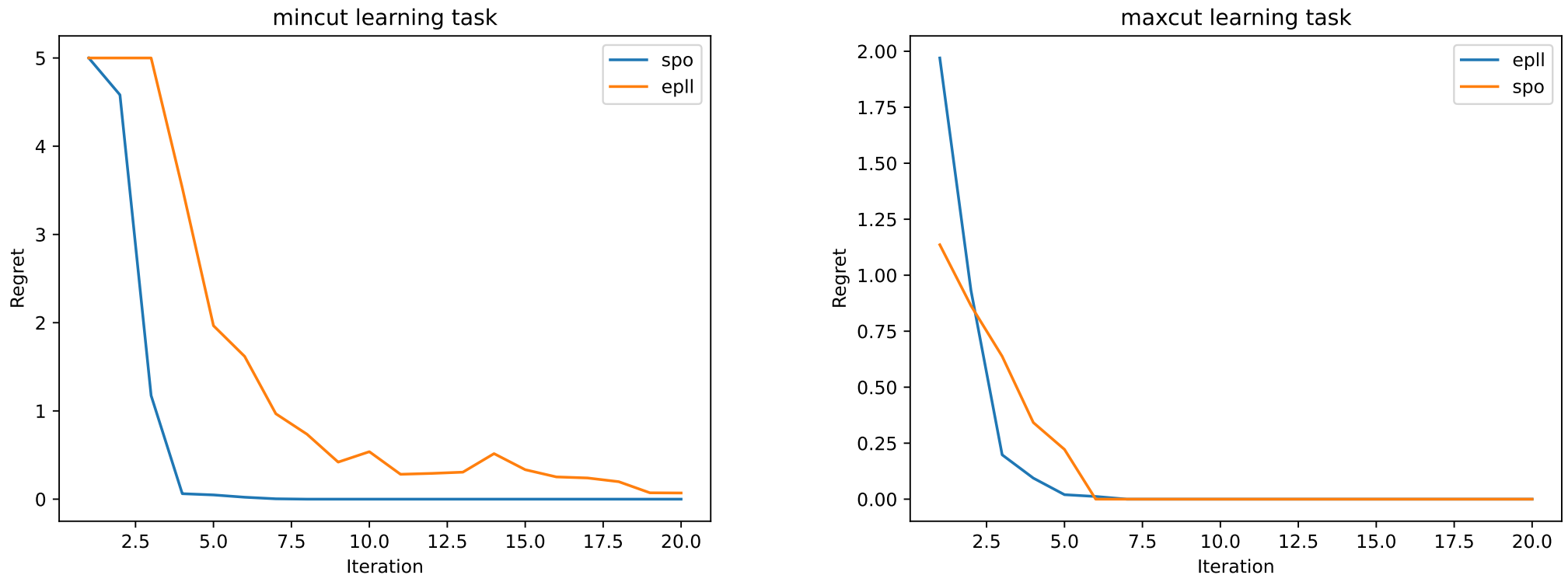

Fig. 5. Test regret vs number of iterations on Min-Cut (left) and Max-Cut (right) tasks, averaged over 10 runs (different initializations).

We adapted the CNN used in task 5.2 to output a single scalar capacity $\hat{c}$ for the input image. This scalar $\hat{c}$ is used as the matrix multiplicative factor to define the GM that will be used, together with $\mathbf{y}^\ell$, to compute the E-PLL loss. The model is trained for 10 epochs over 50 instances using $k = 10$. The training dataset is augmented by duplicating each instance and inverting source and sink nodes (*i.e.,* variables assigned to 1 become 0 and vice-versa). The model is tested every 25 training samples (*i.e.,* quarter of an epoch) on 50 other instances.

We compared the E-PLL with the pioneer DFL SPO+ loss (Elmachtoub and Grigas 2022), both trained with a learning rate of $10^{-3}$. SPO+ requires the true solution $\mathbf{y}^\ell$ for each instance, so it does not save the time needed to convert the training set from a DFL to an SOP setting. Instead, it requires one extra optimisation using $2\hat{\mathbf{c}} - \mathbf{c}$ as the objective function, for every sample during training. On larger graphs, SPO+ training time may become excruciatingly slow.

As shown in Figure 5, both the E-PLL and SPO+ minimize test regret over training iterations. On the Max-Cut task, the E-PLL converges in fewer iterations than SPO+, but it needs more for Min-Cut. This is quite remarkable given that 1) the E-PLL uses a weaker training signal $(y^\ell)$ than SPO+, which exploits both $c^\ell$ and $y^\ell$, and 2) that it also does not explicitly minimize regret. Deciding which approach to use in practice also depends on the problem: some problems may be easier to model and may be more efficiently solved using (I)LP – as done with SPO+ – or using pairwise GMs (Hurley et al. 2016).

While this sub-section is focused on a DFL setting, where constraints are known, our architecture can simultaneously learn the constraints and the objective. The neural net needs just to map each edge image to a $2 \times 2$ matrix (instead of a scalar value), defining the corresponding cost function. The rest of the method is unchanged. To the best of our knowledge, no existing DFL method applies in this complex scenario, where the set of constraints is unknown and needs to be learned.

## 5.4 Learning to Design Proteins

*5.4.1 Problem Definition.* The problem of designing proteins has similarities with solving Sudokus (Strokach et al. 2020). Proteins are linear macro-molecules defined by a sequence of smaller proteins called amino acids. Proteins usually fold into a specific 3D structure which enables specific biological or biochemical functions. Designed proteins have applications in health and green chemistry, among others (Kuhlman and Bradley 2019). To design new proteins, with new functions or enhanced properties, one usually starts from an input backbone structure, matching the target functions, and predicts a sequence of amino acids that will fold onto the target structure.

Considering an input protein structure as a Sudoku grid, each amino acid corresponds to a cell and must be chosen among the 20 natural amino acids instead of 9 digits. The structure is predominantly determined by inter-atomic forces, which drive folding into a minimum energy geometry: the most usual approach of the protein design problem is as an NP-hard unconstrained energy minimization problem (Allouche et al. 2014; Pierce and Winfree 2002). This problem is defined using scoring functions, such as in the widely-used Rosetta (Park et al. 2016), to approximate the laws of physics describing the interactions between amino acids. Inter-atomic forces are influenced by relative distances and atomic natures, which implies that the final interactions inside a protein depend on the geometry of the input structure: we will therefore use this geometry in the natural inputs $\omega$, as we did with the (fixed) Sudoku grid geometry.

Compared to Sudoku, protein design raises several challenges: 1) instances have variable geometries, which is tackled by a dedicated neural network. 2) They may contain several hundred amino acids (*i.e,* variables) or more, requiring an approach able to scale to large instances. 3) No ground-truth cost function is available. In fact, proteins are subjected to many-body interactions that cannot be directly represented in a pairwise GM. We aim to learn a better approximation than the score functions used for design, such as in Rosetta . 4) Many sequences can fit the same input backbone structure, similarly to the one-of-many solutions setting. The parallel between Sudoku and protein design is illustrated in Figure 2.

*5.4.2 Training and Evaluation.* To train the neural network, we use the data set of (Ingraham et al. 2019), already split into train/validation/test sets of respectively $17,000$, $600$, and $1,200$ proteins, in such a way that proteins with similar structures or sequences are in the same set. Similarly to Sudoku, a protein is described by features $\omega$ computed on each pair of positions $i, j$ in the protein. They include inter-atomic distance features encoded with Gaussian radial basis function (Dauparas et al. 2022), and a positional encoding of the sequence distance ($|i-j|$) (Vaswani et al. 2017). Each backbone geometry $\omega$ is associated with $\mathbf{y}$, the sequence of the corresponding known protein. All pair features in $\omega$ are processed by a neural network composed of a recurrent gated MLP (H. Liu et al. 2021) that learns an environment embedding from a central amino acid and its neighbours within 10Å, fed to a ResMLP (as for Sudoku) that takes pairs of features and environment embeddings to predict $20 \times 20$ cost matrices. Training and architecture details are in Annex C.3.

We train the same model with the same initialization using either the NPLL or the E-PLL loss. To adapt to variable protein sizes, the E-PLL eliminates $k$% of incoming neighbor messages. With up to 500 amino acids, the optimisation task is challenging at inference and we used a recent GM convex solver (Durante et al. 2022). One usual metric for protein design is the Hamming distance between the predicted and observed (native) sequences, called the Native Sequence Recovery rate (NSR). Protein design is a multiple-solution problem: above 30% of similarity, two protein sequences are considered as having the same fold (geometry). So, one given structure can adopt many sequences and a 100% NSR cannot be reached.

As shown in Table 4, we observe that the E-PLL not only preserves the good properties of the NPLL but actually improves the NSR. While protein design is often stated as an unconstrained optimisation problem, we hypothesize that this improvement results from the existence of infeasibilities: when local environments are very tight, they absolutely forbid large amino acids. Such infeasibilities could not be properly estimated by the NPLL alone.

Table 4. Comparing the E-PLL and the regular NPLL on the test proteins. Median NSR over the full test set are given.

| Loss | NPLL | **E-PLL** |
|---|---|---|
| NSR | 45.1% | **48.4%** |

Our architecture provides a decomposable scoring function, such as those used for protein design in Rosetta (Park et al. 2016). We compared both approaches on the dataset of (Park et al. 2016) in Table 5. The dataset is restricted to small proteins because designing with Rosetta on larger proteins is intractable. Note that small proteins tend to be less constrained than larger ones, hence the decrease in NSR (Ingraham et al. 2019) compared to Table 4. The cost functions learned with our E-PLL outperforms Rosetta's energy function, leading to designed proteins more similar to natural ones. This is remarkable as Rosetta's full-atom function considers all atoms of the protein while we just use the backbone geometry and amino acids identities (as in coarse-grained scoring functions (Kmiecik et al. 2016)).

Table 5. Comparison with the energy-based design method Rosetta on small single-chain proteins.

| | Rosetta[1] | **E-PLL** |
|---|---|---|
| NSR | 17.9% | **33.0%** |

## 6 Discussion

The results we obtained on all variants of the Sudoku-related tasks show impressive data-efficiency compared to pure DL approaches, and it is tempting to try to explain this difference. The first reason that may explain this data efficiency lies in its inductive biases:

- We focus on learning pairwise interactions and these are exactly those that are present in Sudoku.
- We take into account the geometry of the grid (cell coordinates being provided as input to the network) and this is crucial for the existence of Sudoku constraints.

However, the first bias is also present in all pure DL-based Transformer-based architectures, as attention matrices are precisely aimed at detecting the same pairwise interactions as in pairwise Graphical Models (Bhattacharya et al. 2021). This is also the case for GNN (or RRN) based approaches that focus on (pairwise) graph edges between variables. The input grid-geometry is also provided in Transformer-based approaches, at least indirectly through each variable ROPE positional encoding (G. Wang et al. 2025).

A more convincing explanation is that learning a compact set of rules should be much simpler than learning a function that maps a partially filled Sudoku grid to a solved grid. Very informally, the space of such functions is huge: considering that an "empty cell" defines an extra state, there are $10^{81}$ possible input grids, and $9^{81}$ possible output grids, thus $90^{81} \approx 10^{158}$ such functions. Instead, the number of pairwise Boolean function networks is $\frac{n(n-1)}{2} \cdot 2^{d^2} = 81 \times 40 \times 2^{81} \approx 10^{27}$, which is amazingly smaller. Moreover, contrarily to the function mapping input grids to solutions, the set of rules should be sparse, just because humans don't like games with complex rules (this also applies to protein design, where a linear number of short-range interactions are known to dominate in physics). In our architecture, L1 regularization directs training towards predicting sparse sets of rules. Eventually, the strategy of learning rules and then exploiting them to find solutions seems therefore far more attractive than directly trying to learn a mapping from input grids to solutions.

[1]Rosetta's results are extracted from (Ingraham et al. 2019)

Existing pure-DL approaches based on GNNs and Transformers likely follow this path too, but they face a double challenge:

- They have to learn the correct rules
- They have to learn how to use them (how to reason on them, using, *e.g.*, Message Passing in GNNs)

We suspect that pure DL architectures struggle on the second challenge. Sudoku is NP-complete and extremely hard grids require more compute than what usual DL models can spend. Iterative denoising diffusion probabilistic models and related hierarchical Transformer-based models (G. Wang et al. 2025; Ye et al. 2025) can spend more compute at inference and show better results on Sudoku than direct GNN or Transformer-based models (Palm et al. 2018; Yang et al. 2023), but they are unable to compete with SOTA symbolic solvers that went through decades of improvements. Although deep DL architectures can emulate dynamic programing (and make it differentiable (Mensch and Blondel 2018)), strongly NP-complete problem solving may either require worst-case exponential space/time dynamic programming (as resolution does in logic (Dechter and Rish 1994)) or worst-case polyspace/exponential time tree-search (M. Cooper et al. 2020). SOTA symbolic solvers exploit complex combinations of both (M. Cooper et al. 2020; Marques-Silva et al. 2009). In contrast, usual DL architectures use fixed space/time or linear (quadratic with Transformers) space/time in denoising/recurrent architectures. Finally, it is not clear how DL architectures can emulate "backtracking". Even if LLM "reasoning models" could, in principle, do this through their "reasoning trace", this may require an exponentially long context (something that only very recent Markovian LLM (Aghajohari et al. 2025) may be able to deal with, considering backtracking requires linear space).

In some sense, the situation here is closely related to symbolic regression (Makke and Chawla 2024) or hybrid genetic programming/LLM FunSearch-like approaches (Romera-Paredes et al. 2024): they learn to generate formulae or Python codes that compute a function rather than learning the function directly. In these cases, one can assume the code to be small, which helps learning. Such approaches, however, require assessing the code output during training, which, for NP-complete problem solving, will eventually become excruciating. Our hybrid architecture and its dedicated loss avoid this cost.

## 7 Conclusion

In this paper, we introduced a neuro-symbolic architecture and a dedicated loss function for learning how to solve discrete reasoning and optimisation problems. The architecture is differentiable and as such, allows natural inputs to participate in the definition of discrete reasoning/optimisation problems, providing the ability to inject suitable inductive biases that can also enhance data efficiency, as shown in the case of Sudoku. While optimisation layers (Pogančić et al. 2020; Sahoo et al. 2023; P.-W. Wang et al. 2019) can be inserted in an arbitrary position in a neural net, our neuro-symbolic layer with the E-PLL loss must be the final layer. This restriction is of no consequence in many practical settings, where the discrete decision is the final expected output. It instead offers scalable training, avoiding calls to exact solvers that quickly struggle with the noisy instances that are predicted in early training epochs. It is able to benefit from exact or relaxed solvers during inference. Thanks to the E-PLL, it can simultaneously identify an objective to optimize and constraints to satisfy. Finally, its output can be scrutinized to check properties and can be *a posteriori* completed with side constraints or additional objectives, to inject further instance-dependent information, that may have been independently learned, be available as knowledge, or as user requirements. These facilities are crucially needed in applicative contexts such as protein design (Albanese et al. 2025).

Much remains to be done around this architecture. As it has been done for SAT-Net (Lim et al. 2022), the ultimate $N(\omega)$ GM layer of our architecture could be analyzed during training to identify emerging hypothetical global properties such as symmetries or global decomposable constraints, allowing for more interpretable results.

For memory and computational efficiency reasons, we restricted ourselves to *pairwise* GMs, limiting the detection of complex many-body interactions. The ability of the architecture to deal with latent variables could be combined with dual/hidden representations of constraints (Stergiou and Walsh 1999) to enhance its capacity to fully capture such interactions.

The language of NP-hard discrete Graphical Models with latent variables that we use as the output of our architecture is powerful. Yet, like any NP-hard modeling language, it is not necessarily optimal for concisely representing arbitrary decision NP-complete problems. To extend the range of practical problems that can be efficiently and concisely learned by our architecture, the E-PLL loss should be extended with the ability to deal with extended languages, including, *e.g.*, clauses, linear constraints, and other global constraints and cost functions (Bessiere et al. 2023).

## Acknowledgments

We thank S. Buchet and H. Bierlee for proofreading. Most of this work was done during MD's PhD at INSA Toulouse and ANITI, INRAE. This work has been financially supported by the French national research project GMLaS under grant agreement ANR-24-CE23-3429 and has been supported by the AI-cluster ANITI. ANITI is funded by the France 2030 program under the Grant agreement ANR-23-IACL-0002.

## A Gradient of the NPLL

We have a data set $\mathbf{S}$ composed of $m$ pairs $(\omega^\ell, \mathbf{y}^\ell)$, $1 \leq \ell \leq m$. The Negative Pseudologlikelihood of $\mathbf{S}$ is the sum of the negative log-probability of each $(\omega^\ell, \mathbf{y}^\ell)$:

$$NPLL(\mathbf{S}) = \sum_{\ell=1}^{m} NPLL(\omega^\ell, \mathbf{y}^\ell) = -\sum_{\ell=1}^{m} \left[ \sum_{Y_i \in \mathbf{Y}} \log P^{N(\omega)}(y_i^\ell | \mathbf{y}_{-i}^\ell) \right] \tag{1}$$

where

$$P^{N(\omega)}(y_i^\ell | \mathbf{y}_{-i}^\ell) = \frac{\exp(-\sum_{j \neq i} N(\omega)[i,j](y_i^\ell, y_j^\ell))}{\sum_{v_i \in D^i} \exp(-\sum_{j \neq i} N(\omega)[i,j](v_i, y_j^\ell))}$$

The conditional probability above is obtained using the normalizing constant $Z^{N(\omega)}(\mathbf{y}_{-i}^\ell)$ in the denominator:

$$Z^{N(\omega)}(\mathbf{y}_{-i}^\ell) = \sum_{v_i \in D^i} \exp(-\sum_{j \neq i} N(\omega)[i,j](v_i, y_j^\ell))$$

computed over all possible values $v_i$ of $Y_i$.

Minimizing the NPLL means maximizing the probability above, therefore making $-N(\omega)[i,j](\cdot, y_j^\ell)$ higher on the observed value $y_i^\ell$ (used in the numerator) than on the other values $v_i \neq y_i^\ell$ or equivalently, the cost $N(\omega)[i,j](\cdot, y_j^\ell)$ lower on $y_i^\ell$ than on other values: the NPLL is a contrastive loss that seeks to create a margin between the values that are observed in the sample $S$ and the other values of the variable, for every variable and every sample.

Focusing on one pair $(\omega, \mathbf{y}) \in S$, we expand and get:

$$NPLL(\omega, \mathbf{y}) = -\sum_{Y_i \in \mathbf{Y}} \left[ \left( -\sum_{j \neq i} N(\omega)[i,j](y_i, y_j) \right) - \log Z^{N(\omega)}(\mathbf{y}_{-i}) \right] \tag{2}$$

The NPLL is a sum over all variables $Y_i \in \mathbf{Y}$. We consider the contribution of a given variable $Y_i$. To compute the gradients of the corresponding term of the NPLL, we first compute the partial derivative of the logarithm of the normalizing constant $Z^{N(\omega)}(\mathbf{y}_{-i})$ ($i$ fixed) w.r.t. $N(\omega)[i,j](v_i, y_j)$ (for arbitrary $j \neq i$ and $v_i \in D^i$, other costs do not appear in $Z^{N(\omega)}(\mathbf{y}_{-i})$ and the corresponding partial derivative is 0).

$$\frac{\partial \log Z^{N(\omega)}(\mathbf{y}_{-i})}{\partial N(\omega)[i,j](v_i, y_j)} = \frac{-\exp(-\sum_{k \neq i} N(\omega)[i,k](v_i, y_k))}{Z^{N(\omega)}(\mathbf{y}_{-i})} \qquad = -P^{N(\omega)}(v_i | \mathbf{y}_{-i})$$

For any $Y_i$, the partial derivative of the first term in equation 2 w.r.t. $N(\omega)[i,j](v_i,v_j)$ is $-\mathbb{1}(v_i = y_i, v_j = y_j)$. Overall, given that $N(\omega)[i,j](v_i,v_j)$ and $N(\omega)[j,i](v_j,v_i)$ are the same, the contribution of sample $(\omega,\mathbf{y})$ to $\frac{\partial NPLL}{\partial N(\omega)[i,j](v_i,v_j)}$ will reduce to the non-zero contributions of variables $Y_i$ and $Y_j$:

$$\begin{aligned}\frac{\partial NPLL}{\partial N(\omega)[i,j](v_i,v_j)} &= [\mathbb{1}(y_i = v_i, y_j = v_j) - P^{N(\omega)}(v_i|\mathbf{y}_{-i})\mathbb{1}(y_j = v_j)] \\ &\quad + [\mathbb{1}(y_i = v_i, y_j = v_j) - P^{N(\omega)}(v_j|\mathbf{y}_{-j})\mathbb{1}(y_i = v_i)] \end{aligned} \tag{3}$$

# B Theoretical Analysis of the Negative Pseudo Log-Likelihood (NPLL)

## B.1 The Convex NPLL is Asymptotically Consistent

In this section, we assume that a dataset $\mathbf{S}$ of $m$ samples $(y^\ell)$ has been generated by a Markov Random Field $\mathcal{N}^*$. As above, we will write the NPLL and its gradient in terms of the parameters of a currently estimated MRF with parameters $\mathcal{N}_{ij}(a,b)$. By bringing the sum over samples $\ell$ inside, we will make the frequencies (probabilities) of tuples of values in the dataset $\mathbf{S}$ visible. We consider the average NPLL over $\mathbf{S}$, defined as:

$$\overline{NPLL}(\mathbf{S}) = -\frac{1}{m}\sum_{\ell=1}^{m}\left[\sum_{Y_i\in\mathbf{Y}} \log P^{\mathcal{N}}(y_i^\ell|\mathbf{y}_{-i}^\ell)\right]$$

where

$$P^{\mathcal{N}}(y_i^\ell|\mathbf{y}_{-i}^\ell) = \frac{\exp(-\sum_{j\neq i}\mathcal{N}_{ij}(y_i^\ell, y_j^\ell))}{\sum_{v_i\in D^i}\exp(-\sum_{j\neq i}\mathcal{N}_{ij}(v_i, y_j^\ell))}$$

The conditional probability above is obtained using the normalizing constant $Z^{\mathcal{N}}(\mathbf{y}_{-i}^\ell)$ in the denominator:

$$Z^{\mathcal{N}}(\mathbf{y}_{-i}^\ell) = \sum_{v_i\in D^i}\exp(-\sum_{j\neq i}\mathcal{N}_{ij}(v_i, y_j^\ell)) \tag{4}$$

computed over all possible values $v_i$ of $Y_i$. This gives:

$$\begin{aligned}\overline{NPLL}(\mathbf{S}) &= \frac{1}{m}\sum_{\ell=1}^{m}\sum_{Y_i\in\mathbf{Y}}\sum_{j\neq i}\mathcal{N}_{ij}(y_i^\ell, y_j^\ell) + \sum_{\ell=1}^{m}\sum_{Y_i\in\mathbf{Y}}\log Z^{\mathcal{N}}(\mathbf{y}_{-i}^\ell) \\ &= \frac{1}{m}\sum_{Y_i\in\mathbf{Y}}\sum_{j\neq i}\sum_{\ell=1}^{m}\mathcal{N}_{ij}(y_i^\ell, y_j^\ell) + \sum_{Y_i\in\mathbf{Y}}\sum_{\ell=1}^{m}\log Z^{\mathcal{N}}(\mathbf{y}_{-i}^\ell) \\ &= \sum_{Y_i\in\mathbf{Y}}\sum_{j\neq i}P_{\mathbf{S}}(y_i, y_j)\mathcal{N}_{ij}(y_i, y_j) + \sum_{\mathbf{y}_{-i}}P_{\mathbf{S}}(\mathbf{y}_{-i})\log Z^{\mathcal{N}}(\mathbf{y}_{-i})\end{aligned}$$

where $P_{\mathbf{S}}$ is the probability in the dataset $\mathbf{S}$ and $P_{\mathbf{S}}(y_i, y_j)\mathcal{N}_{ij}(y_i, y_j)$ and $P_{\mathbf{S}}(y_{-i})\log Z^{\mathcal{N}}(\mathbf{y}_{-i})$ are summed over all non-zero probabilities. At this point, it is interesting to note that the NPLL is convex, as the sum of linear and convex functions (LogSumExp of linear functions).

Following the same derivation as above for the gradient, we have:

$$\frac{\partial \log Z^{\mathcal{N}}(\mathbf{y}_{-i})}{\partial \mathcal{N}_{ij}(y_i, y_j)} = \frac{-\exp(-\sum_{k\neq i} \mathcal{N}_{ik}(y_i, y_k))}{Z^{\mathcal{N}}(\mathbf{y}_{-i})} = -P^{\mathcal{N}}(y_i|\mathbf{y}_{-i}) \tag{5}$$

The gradient of $\overline{NPLL}$ with respect to a given parameter $\mathcal{N}_{ij}(y_i, y_j)$ (fixing $y_i$ and $y_j$)

$$\begin{aligned}\frac{\partial \overline{NPLL}}{\partial \mathcal{N}_{ij}(y_i, y_j)} &= (P_{\mathbf{S}}(y_i, y_j) - \sum_{\mathbf{y}_{-i-j}} P_{\mathbf{S}}(\mathbf{y}_{-i})P^{\mathcal{N}}(y_i|\mathbf{y}_{-i})) \\ &+ (P_{\mathbf{S}}(y_i, y_j) - \sum_{\mathbf{y}_{-j-i}} P_{\mathbf{S}}(\mathbf{y}_{-j})P^{\mathcal{N}}(y_j|\mathbf{y}_{-j}))\end{aligned}$$

As $m \to \infty$, we have $P_{\mathbf{S}}(\mathbf{y}) \to P^{\mathcal{N}^*}(\mathbf{y})$ and the gradient becomes:

$$(P^{\mathcal{N}^*}(y_i, y_j) - \sum_{\mathbf{y}_{-i-j}} P^{\mathcal{N}^*}(\mathbf{y}_{-i})P^{\mathcal{N}}(y_i|\mathbf{y}_{-i})) + (P^{\mathcal{N}^*}(y_i, y_j) - \sum_{\mathbf{y}_{-j-i}} P^{\mathcal{N}^*}(\mathbf{y}_{-j})P^{\mathcal{N}}(y_j|\mathbf{y}_{-j}))$$

At the true value $\mathcal{N}^*$ of $\mathcal{N}$, this becomes equal to:

$$\begin{aligned}&(P^{\mathcal{N}^*}(y_i, y_j) - \sum_{\mathbf{y}_{-i-j}} P^{\mathcal{N}^*}(\mathbf{y}_{-i})P^{\mathcal{N}^*}(y_i|\mathbf{y}_{-i})) + (P^{\mathcal{N}^*}(y_i, y_j) - \sum_{\mathbf{y}_{-j-i}} P^{\mathcal{N}^*}(\mathbf{y}_{-j})P^{\mathcal{N}^*}(y_j|\mathbf{y}_{-j})) \\ = &(P^{\mathcal{N}^*}(y_i, y_j) - \sum_{\mathbf{y}_{-i-j}} P^{\mathcal{N}^*}(\mathbf{y}_{-i})P^{\mathcal{N}^*}(y_i|\mathbf{y}_{-i})) + (P^{\mathcal{N}^*}(y_i, y_j) - \sum_{\mathbf{y}_{-j-i}} P^{\mathcal{N}^*}(\mathbf{y}_{-j})P^{\mathcal{N}^*}(y_j|\mathbf{y}_{-j}))\end{aligned}$$

Consider the second term above:

$$\sum_{\mathbf{y}_{-i-j}} P^{\mathcal{N}^*}(\mathbf{y}_{-i})P^{\mathcal{N}^*}(y_i|\mathbf{y}_{-i}) = \sum_{\mathbf{y}_{-i-j}} P^{\mathcal{N}^*}(y_i, \mathbf{y}_{-i}) = P^{\mathcal{N}^*}(y_i, y_j)$$

which also applies to the last term, leading to:

$$\frac{\partial \overline{NPLL}}{\partial \mathcal{N}_{ij}(y_i, y_j)} = (P^{\mathcal{N}^*}(y_i, y_j) - P^{\mathcal{N}^*}(y_i, y_j)) + (P^{\mathcal{N}^*}(y_i, y_j) - P^{\mathcal{N}^*}(y_i, y_j)) = 0$$

The NPLL is therefore convex and asymptotically reaches a critical point on the true values. Under the assumption of strict convexity, this means that the NPLL reaches its unique minimum at the true values. In practice, as in the protein design case, data is rarely generated by a pairwise MRF and may include more complex interactions. The existence of zero probabilities also creates obvious non-identifiabilities because of the existence of redundant constraints.

## B.2 Analysing the PLL as a Fenchel-Young Loss

Fenchel-Young Losses (Blondel et al. 2020) (FYL) are losses that are built based on a regularized prediction function. In the context of SOP, when the structure $\mathcal{Y}$ is captured by a polytope and the objective function is linear, one vertex of the polytope will be an optimal solution. During learning, the learned objective parameters $\mathbf{c}$ evolve, and this leads to sudden shifts of the optimal discrete solution to a new vertex. Regularization in the convex hull $conv(\mathcal{Y})$ of the solution space $\mathcal{Y}$ provides a smoothing component that creates differentiability.

Denoting $\Omega(\mu)$ the regularization function, $\mu \in conv(\mathcal{Y})$, the regularized prediction function is defined as $\hat{\mathbf{y}}_\Omega(\mathbf{c}) \in \arg\max_{\mu \in conv(Y)}(\langle \mathbf{c}, \mu\rangle - \Omega(\mu))$ and the associated Fenchel-Young loss will be $\Omega(\mathbf{y}) - \langle \mathbf{c}, \mathbf{y}\rangle + \Omega^*(\mathbf{c})$ where $\Omega^*(\mathbf{c})$ is the Fenchel-Young conjugate of the regularization function $\Omega(\mathbf{y})$ (also called the potential function of the loss). The regularized prediction function $\hat{\mathbf{y}}_\Omega(\mathbf{c})$ resides in the convex hull $conv(\mathcal{Y})$ and can therefore be seen as a convex combination of elements of $\mathcal{Y}$.

For pairwise MRFs, the sufficient statistics $\phi(\mathbf{y})$ of an assignment $\mathbf{y}$ are defined over pairs of values of pairs of variables $(y_i, y_j)$. With the MRF negative log-likelihood (NLL) loss, also known as the Conditional Random Field (CRF) loss, the potential function is the intractable log-partition function (See (Blondel et al. 2020, p. 10)). For tractability reasons, the NPLL is sparse. Its support includes only a neighborhood $N_1(\mathbf{y}) \subset \mathcal{Y}$ around the true output $\mathbf{y}$, obtained by changing the value of at most one variable:

$$N_1(\mathbf{y}) = \{\mathbf{y'} \in \mathcal{Y} \mid \text{Hamming}(\mathbf{y'}, \mathbf{y}) \leq 1\}$$

The structure of the NPLL suggests using a potential function $\Omega^*_{PLL}(\theta; \mathbf{y}) = \frac{1}{n}\sum_{i=1}^{n} \log Z^\theta(\mathbf{y}_{-i})$ (see eq. 4). Then, $\Omega_{PLL}(\mu; \mathbf{y}) = \sup_\theta\{\langle\theta, \mu\rangle - \frac{1}{n}\sum_{i=1}^{n} Z^\theta(\mathbf{y}_{-i})\}$. Let $\theta^*(\mu, \mathbf{y})$ be the value of $\theta$ that achieves the supremum. This means that $\mu = \frac{1}{n}\sum_{i=1}^{n} \nabla_\theta Z^\theta(\mathbf{y}_{-i})$. Using eq. 5, $\mu$ is a sum of expected sufficient statistics $\mathbb{E}_{P^\theta(y_i'|\mathbf{y}_{-i})}[\phi((y_i', \mathbf{y}_{-i}))]\rangle$. The conditional entropy of the distribution $P^{\theta^*(\mu,\mathbf{y})}(Y_i'|\mathbf{y}_{-i})$ is $H^\theta(Y_i|\mathbf{y}_{-i}) = -\sum_{y_i'} P^\theta(y_i'|\mathbf{y}_{-i}) \log P^\theta(y_i'|\mathbf{y}_{-i})$. Now, $\log P^\theta(y_i'|\mathbf{y}_{-i}) = \langle\theta, \phi((y_i', \mathbf{y}_{-i}))\rangle - Z^\theta(\mathbf{y}_{-i})$ and $H^\theta(Y_i|\mathbf{y}_{-i}) = -\sum_{y_i'} P^\theta(y_i'|\mathbf{y}_{-i})(\langle\theta, \phi(y_i', \mathbf{y}_{-i})\rangle - \log Z^\theta(\mathbf{y}_{-i})) = \log Z^\theta(\mathbf{y}_{-i}) - \langle\theta, \mathbb{E}_{P^\theta(y_i'|\mathbf{y}_{-i})}[\phi((y_i', \mathbf{y}_{-i}))]\rangle$. Therefore, $\Omega_{PLL}(\mu; \mathbf{y}) = -\sum_{i=1}^{n} H^{\theta^*(\mu;\mathbf{y})}(Y_i|\mathbf{y}_{-i})$.

This shows how the NPLL can be interpreted as a Fenchel-Young loss, regularized by a short list of neighbor assignments weighted by local conditional probabilities. However, this Fenchel-Young view relies on a potential function $\Omega^*$ and a regularization function $\Omega$ that both depend on $\mathbf{y}$. This differs from most Fenchel-Young losses, such as the CRF log-likelihood, the structured perception, or the SparseMAP losses. It shares this dependency on $\mathbf{y}$ with the structured hinge loss, where $\theta$ alone is modified using a cost vector that depends on $\mathbf{y}$.

## C Experiment Details

### C.1 Generation of Futoshiki Dataset

Grids with 5 rows and 5 columns were created. Starting from an empty grid, inequalities were generated randomly. For each pair of adjacent cells, an inequality was added with a fixed uniform probability. The inequality was turned into a constraint in toulbar2, then the solver was asked to solve the grid. One grid may have several solutions. To avoid the introduction of bias by selecting one, random unary costs drawn from a uniform distribution are added to each variable. This way, a small random cost is added to each solution, helping toulbar2 to select one uniformly. This process was repeated to generate the entire dataset.

### C.2 Robustness to Parameter $k$

On the symbolic Sudoku task, we assess the impact of the parameter $k$ of the Emmental-NPLL, representing the number of holes. A value of $k = 0$ corresponds to the regular NPLL, with which the neural net fails to solve any test grids. Otherwise, as displayed in Table 6, the E-PLL is robust to most values of $k$, as long as some variables are masked ($k > 0$) and some are not masked ($k < 80$). These results were obtained with a larger neural net (10 hidden layers with 128 neurons each) than experiments from Subsection 5.1.

### C.3 Architecture and Training Details

For the protein design task, the neural network is composed of an input linear layer of size 128, a block to extract environment information, and a ResMLP to predict cost matrices. The environment network is composed of a gated MLP and a ResNet, repeated 6 times (with shared weights), the output of the previous iteration being added to the input. The gatedMLP has 3 layers, a width of 2 × 256 and it considers the 48 nearest neighbours of each residue. All ResMLP contains 6 layers, with a recurrent connection every 2 layers, and a width of 256. The resulting architecture has 3.4M parameters.

Table 6. Average performances over 10 initializations, for various values of parameter $k$. Training is up to 100 epochs.

| $k$ | Epochs | Training time ($s$) | Runs with 100% test grids solved |
|---|---|---|---|
| 0 | 100 | - | 0% |
| 10 | $23.2 \pm 2.6$ | $566 \pm 67$ | 100% |
| 20 | $38.6 \pm 6.9$ | $900 \pm 151$ | 90% |
| 50 | $50.4 \pm 7.6$ | $1257 \pm 184$ | 90% |
| 70 | $27.2 \pm 2.7$ | $724 \pm 83$ | 100% |
| 80 | 100 | - | 0% |

As in most protein energy functions (Alford et al. 2017), amino acid pairs separated by a large distance are ignored. We use a 15 Å threshold. The neural network is trained using the Adam optimizer, with a weight decay of $10^{-3}$ and an initial learning rate of $5.10^{-4}$, divided by 10 when the validation loss decreases (with patience 0) until it reaches $10^{-*}$. A L1 regularization of $10^{-4}$ is applied to costs. To compare the NPLL loss and the E-PLL loss, we trained the same model with the same hyperparameters and starting from the same weight initialization with each of the loss. Both models run as long as the minimum LR is not reached, and therefore not necessarily for the same number of epochs.